\documentclass[runningheads]{llncs}
\usepackage[T1]{fontenc}
\usepackage{graphicx}
\usepackage{verbatim}
\usepackage{hyperref}
\usepackage{color}

\usepackage[table,xcdraw]{xcolor} 
\usepackage{amsmath}
\usepackage{amsfonts}  
\usepackage{bm}
\usepackage{bbm}  
\usepackage{array}  
\usepackage{booktabs}  
\usepackage{multirow}  
\usepackage{makecell}  
\usepackage{caption}
\usepackage{subcaption}
\usepackage{bbding}  
\usepackage{soul}    
\usepackage{pifont}  
\usepackage{upgreek}
\newcommand{\rc}[1]{REAL-Colon}
\newcommand{\sun}[1]{SUN database}
\newcommand{\pset}[1]{PolypsSet}
\newcommand{\psize}[1]{PolypSize}
\newcommand{\ReID}[1]{ReID}
\newcommand{\reid}[1]{re-identification}
\newcommand{\sota}[1]{state-of-the-art}
\newcommand{\cmark}{\ding{51}}%
\newcommand{\xmark}{\ding{55}}%
\begin{document}
\title{Contrastive Learning under Noisy Temporal Self-Supervision for Colonoscopy Videos}
\titlerunning{Contrastive Learning under Noisy Temporal Self-Supervision in Colonoscopy}
\author{
Luca Parolari\index{Parolari, Luca}\inst{1,2}\Envelope
\and Pietro Gori\index{Gori, Pietro}\inst{2}
\and Lamberto Ballan\index{Ballan, Lamberto}\inst{1}
\and
\\
Carlo Biffi\index{Biffi, Carlo}\inst{3}\thanks{Shared senior authorship.}
\and Loic Le Folgoc\index{Le Folgoc, Loic}\inst{2}\protect\footnotemark[1]
}
\authorrunning{L. Parolari et al.}
\institute{
Department of Mathematics, University of Padova, Padova, Italy
\email{luca.parolari@phd.unipd.it}
\and LTCI, Telecom Paris, Institut Polytechnique de Paris, Palaiseau, France
\and Cosmo Intelligent Medical Devices, Dublin, Ireland 
}
\maketitle              
\begin{abstract}

Learning robust representations of polyp tracklets is key to enabling multiple AI-assisted colonoscopy applications, from polyp characterization to automated reporting and retrieval.
Supervised contrastive learning is an effective approach for learning such representations, but it typically relies on correct positive and negative definitions.
Collecting these labels requires linking tracklets that depict the same underlying polyp entity throughout the video, which is costly and demands specialized clinical expertise.
In this work, we leverage the sequential workflow of colonoscopy procedures to derive self-supervised associations from temporal structure.
Since temporally derived associations are not guaranteed to be correct, we introduce a noise-aware contrastive loss to account for noisy associations.
We demonstrate the effectiveness of the learned representations across multiple downstream tasks, including polyp retrieval and re-identification, size estimation, and histology classification.
Our method outperforms prior self-supervised and supervised baselines, and matches or exceeds recent foundation models across all tasks, using a lightweight encoder trained on only 27 videos.
Code is available at \href{https://github.com/lparolari/ntssl}{github.com/lparolari/ntssl}.

\keywords{Contrastive Learning \and Representation Learning \and Temporal Self-Supervision \and Colonoscopy Videos.}

\end{abstract}
\section{Introduction} 

Colonoscopy is the gold standard for colorectal cancer prevention. 
It enables the detection and removal of adenomatous polyps, \textit{i.e.,} abnormal tissue growths in the colon that can become cancerous~\cite{hassan2021performance}. 
Computer-aided systems based on deep learning have demonstrated the potential to improve patient outcomes and the cost-effectiveness of screening~\cite{areia2022cost}.
In colonoscopy, deep learning systems support applications ranging from polyp detection to downstream tasks such as polyp classification and size estimation, and can further assist post-procedure workflows including structured polyp reporting and lesion retrieval~\cite{nogueirarodriguez2021deep,parolari2025temporally,sekiguchi2026artificial,xiang2024vtreid}.
These tasks rely on robust representations of polyp \emph{tracklets}, \textit{i.e.,} sequences of consecutive detections of the same polyp, and typically depend on dense annotations.
Obtaining such supervision requires localizing polyps throughout the procedure, forming tracklets and associating them with the underlying polyp entity.
This process is costly, time-consuming, and needs specialized clinical expertise, limiting the scalability of supervised learning approaches.
Self-supervised learning offers a promising direction to reduce reliance on such labeled data.

Prior work adopts the SimCLR framework~\cite{chen2020simple} to learn polyp tracklet representations by splitting each tracklet into two disjoint segments and enforcing similarity between the resulting positive views~\cite{intrator2023self}.
Subsequent work aggregates tracklets through clustering to form a more stable polyp-level representation, improving robustness in downstream tasks~\cite{parolari2025towards}.
Another approach leverages the momentum-contrast framework~\cite{he2020momentum} and incorporates textual reports to learn multi-modal tracklet representations~\cite{xiang2024vtreid}.
Recent work~\cite{parolari2025temporally} exploits polyp entity annotations, \textit{i.e.}, labels that link detections to the same underlying polyp, to build ground-truth positive associations between tracklets for supervised contrastive learning.
With this explicit supervision, the method achieves superior performance over self-supervised alternatives, indicating a persistent gap between supervised and self-supervised learning.

In this work, we aim to narrow this gap while eliminating the reliance on polyp entity annotations.
Colonoscopy presents a highly challenging visual environment for self-supervised learning. 
Semantic cues are often sparse, and polyp appearance can vary markedly over time due to changes in viewpoint, scale and illumination, tissue deformation, motion blur, occlusions, debris, and the presence of endoscopic instruments. 
Despite these difficulties, we observe that the clinical workflow remains structured and largely sequential.
The endoscopist inspects the colon, removes a polyp when encountered, and then proceeds to the next one. 
We posit that this procedural structure arising from full-procedure videos provides a valuable self-supervisory signal.
Observations captured close in time are likely to depict the same polyp entity. 
Thus, we build associations from temporal structure and employ contrastive learning to learn shared semantic information between such observations. 
However, the induced temporal associations may occasionally be incorrect, for instance when distinct polyps appear within a short time interval, leading to a noisy self-supervision signal.
To address this, we propose a noise-aware loss that mitigates errors from wrong temporal associations, preventing them from degrading the learned representation.

We evaluate the effectiveness of the learned representations on a wide range of downstream tasks, including polyp retrieval and re-identification, size estimation, and histology classification--on open-access datasets including \rc{}~\cite{biffi2024real}, SUN~\cite{misawa2021development}, PolypSize~\cite{song2025polypsize} and PolypsSet~\cite{li2021colonoscopy}.
Our method outperforms prior self-supervised as well as supervised baselines, highlighting the effectiveness of the proposed temporal self-supervision and noise-aware loss.
We also compare against recent general-purpose and endoscopy-specific foundation models and achieve comparable or superior results across all tasks using a lightweight architecture trained on only 27 videos.

In summary, this work focuses on learning robust polyp tracklet representations without relying on manual supervision. 
Our main contributions are threefold.
(i) A self-supervised framework that exploits the sequential workflow of colonoscopy procedures to construct a self-supervision signal from temporal structure. 
(ii) A noise-aware contrastive loss that mitigates the impact of noisy temporal associations. 
(iii) An extensive evaluation on four public datasets, comparing to self-supervised, fully supervised and foundation model baselines.

\section{Method}

\begin{figure}[t]
    \centering
    \includegraphics[width=1\linewidth]{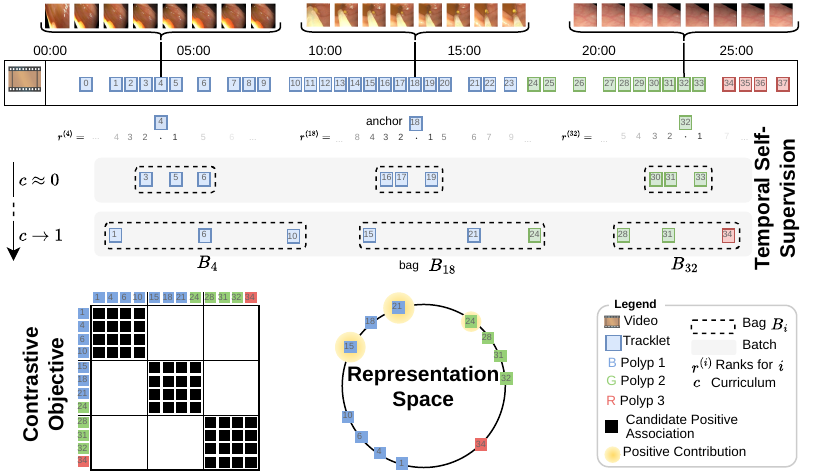}
    \caption{Our method. (Top) Tracklets detected in a colonoscopy video are used to derive the self-supervision and produce bags of tracklets for contrastive learning. At the beginning of training ($c \approx 0$), bags are constituted by temporally adjacent tracklets, yielding low-variation views that are likely to correspond to the same polyp. Later ($c \rightarrow 1$), bags progressively include more distant tracklets, increasing diversity but also the chance of incorrect associations. (Bottom) The model learns from within-bag associations while limiting noisy ones. In the example, the loss maximizes the similarity between tracklet 18, 15 and 21 while down-weighting the contribution of noisy associations such as tracklet 24.}
    \label{fig:method}
\end{figure}

In this section, we present our framework for learning the representation of polyp tracklets under temporal self-supervision (Fig.~\ref{fig:method}).
First, we describe how temporal structure can be used to construct a self-supervision signal.
Then, we introduce our contrastive loss designed to learn from noisy associations.

\subsection{Temporal Structure as Self-Supervision}

Following related works~\cite{intrator2023self,parolari2025towards,parolari2025temporally}, we start by constructing a dataset of $N$ tracklets $\{x_i\}_{i=1}^N$ from ground-truth polyp detections, so that representation learning can be evaluated independently of detection and tracking errors.
A tracklet is a sequence 
of fixed length $L$
constituted by consecutive detections from the same polyp.
Let $v_i$ denote the colonoscopy video containing the $i$-th tracklet and $p_i$ its temporal position within the video. 
Both $v_i$ and $p_i$ are inherently available from the video data and require no manual annotation.
For each tracklet $i$, we define the same-video candidate set $\bm{\mathcal{C}} = \{ j \in \{1, \ldots, N\} \setminus \{i\} \mid v_j = v_i \}$ and compute an index $\pi_i(\cdot)$ to sort the $C$ elements of $\bm{\mathcal{C}}$ by increasing temporal distance from $i$:
$|p_{\pi_i(1)} - p_i| \le |p_{\pi_i(2)} - p_i| \le \dots \le |p_{\pi_i(C)} - p_i|$.
The position in this ordering defines a ranking, where lower ranks correspond to tracklets that are temporally closer to $i$.
Importantly, using ranks lets us define temporal neighborhoods without relying on absolute time windows or hand-tuned thresholds.
\\
For each tracklet $i$, called \textit{anchor}, our goal is to provide a set of candidate positive tracklets, called \textit{bag}, for contrastive learning derived only from temporal associations.

\vspace{0.5em}
\noindent \textbf{Bag Construction}\quad
Given $K$ the bag size, $K \leq C$, a simple strategy is to build each bag by selecting the $K$ samples that are closest in time to the anchor $i$ and build the corresponding bag $B_i = \{x_{\pi_i(1)}, \ldots, x_{\pi_i(K)}\}$.
Although simple, this strategy yields highly redundant bags, making the contrastive task overly easy and limiting the diversity of positive views.
We address this problem by sampling ranks from an exponential distribution. 
Specifically, given an anchor $i$, we sample $K$ distinct ranks $r^{(i)}_\ell \ge 1, \ell \in \{1, \ldots, K\}$ without replacement from:
$
\mathbb{P}_\tau(r)
=
\frac{\exp(-r / \tau)}{\sum_{u=1}^{C} \exp(-u / \tau)}.
$
The resulting bag is
$
B_i = \{x_m \mid m = \pi_i(r^{(i)}_\ell),\ \ell = 1,\dots,K \}.
$
The temperature $\tau$ shifts probability mass from lower to higher ranks, thereby controlling the diversity of tracklets in the bag.
When $\tau \approx 0$, sampling concentrates on tracklets closest to $i$, yielding candidate positives that are highly similar to each other but very likely represent the same polyp. 
As $\tau$ increases, the distribution flattens and increasingly higher ranks are selected, resulting in candidate positives that capture larger temporal and appearance variation, but with a higher risk of including incorrect temporal associations
(Fig.~\ref{fig:method} (Top)).

\begin{figure}[h]
\centering
\begin{subfigure}[t]{0.49\textwidth}
\centering
\includegraphics[width=\linewidth]{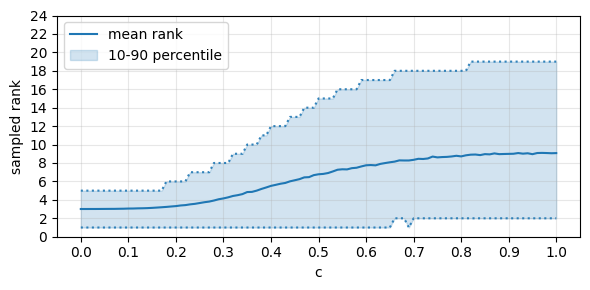}
\end{subfigure}
\hfill
\begin{subfigure}[t]{0.49\textwidth}
\centering
\includegraphics[width=\linewidth]{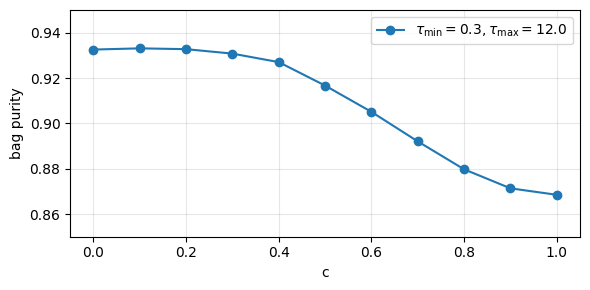}
\end{subfigure}
\caption{(Left) Sampled rank throughout the curriculum. When $c$ increases, higher ranks are sampled. (Right) Ratio of pure bags (\textit{i.e.,} bags containing only tracklets from the same polyp) throughout the curriculum.}
\label{fig:sampler}
\end{figure}

\vspace{0.5em}
\noindent\textbf{Curriculum}\quad
Rather than using a fixed temperature $\tau$, we adopt curriculum learning~\cite{bengio2009curriculum} and gradually increase $\tau$ from low to high values over training.
Specifically, we define a curriculum variable $c \in [0,1]$ that tracks training progress in time, and set $\tau = \mathcal{T}(c)$ using a cosine schedule:
$
\mathcal{T}(c)
=
\tau_{\min}
+
\frac{1-\cos(\pi c)}{2}\,(\tau_{\max}-\tau_{\min}).
$
Fig.~\ref{fig:sampler} (Left) visualizes the ranks sampled throughout the curriculum.

\subsection{Noise-Aware Contrastive Loss}

We consider a dataset of paired samples $\{(B_i, a_i)\}_{i=1}^N$,
where $a_i$ is an anchor tracklet and $B_i = \{b_{i1}, b_{i2}, \ldots, b_{iK}\}$ is a bag of tracklets associated with $a_i$.
The composition of $B_i$ is crucial for effective contrastive representation learning.
Positive samples should be sufficiently similar to the anchor to convey shared semantic content, yet diverse enough to promote robustness to appearance variations.
Although colonoscopy procedures exhibit a sequential structure, temporal proximity does not always provide correct supervision.
During the construction of bags, the sampler may include tracklets whose polyp entity differs from that of the anchor, particularly when higher-rank tracklets are considered.
We refer to such bags as impure.
Impure bags are undesirable because different polyps may exhibit different clinical properties (e.g. size or histology). 
Treating all tracklets in an impure bag as positives can introduce spurious supervision and degrade learning when applied naively within a contrastive framework.
Fig.~\ref{fig:sampler} (Right) illustrates this issue, showing that the proportion of impure bags increases as the curriculum progresses.

In light of this, we impose the following condition on the representation space that we want to learn:
for each anchor $a_i$, at least one tracklet in $B_i$ should be more similar to $a_i$ than any tracklet in $B_i$ is to any other anchor $a_j$.
This condition relaxes traditional contrastive learning constraints where all positive samples contribute equally, such as in  SupCon~\cite{khosla2020supcon}. 
It can be satisfied by just one of the positives, allowing our model to ignore noisy associations and therefore prevent the degradation of the learned space~\cite{miech2020milnce}
(Fig.~\ref{fig:method} (Bottom)).
Formally, let $f_\theta: x \mapsto f_\theta(x) \in \mathbb{R}^d$ be an encoder which maps tracklets into a shared representation space.
Let $z_{ik} = f_\theta(b_{ik})$, $y_i = f_\theta(a_i)$ and $y_j = f_\theta(a_j)$.
Let $s(\cdot, \cdot)$ be a similarity function in the joint embedding space. We can mathematically translate the condition into:
$
\max_k s(z_{ik}, y_i) > \max_k s(z_{ik}, y_j),
$
$\forall j \neq i.$
As commonly done in contrastive learning~\cite{barbano2023unbiased,dufumier_integrating_2023}, this inequality can be transformed into an optimization problem by using the $\max$ operator and its smooth approximation log-sum-exp to derive the loss function:
\begin{equation}
\mathcal{L}
=
-
\sum_i
\log \Bigg( 
\frac{ 
    \sum_{k=1}^{K} \exp \big(s(z_{ik},y_i) \big)
}{
    \sum_{k=1}^{K} \exp \big(  s(z_{ik},y_i) \big) 
    + \sum_{\substack{j=1 \\ j \neq i}}^{N} \sum_{k=1}^{K} \exp \big(  s(z_{ik},y_j) \big)
}  \Bigg)
\end{equation}

\vspace{2ex}
\noindent\textbf{Multi-Level Objective}\quad
The proposed loss is architecture-agnostic. 
To ensure fair comparison with prior work, we adopt the same lightweight tracklet encoder used in related approaches~\cite{intrator2023self,parolari2025towards,parolari2025temporally}. 
It uses a transformer encoder with a learnable token~\cite{devlin2019bert} prepended to the sequence to aggregate a global representation of the tracklet. 
In addition, the encoder produces $L$ embeddings, one per frame, analogous to patch tokens in Vision Transformers~\cite{dosovitskiy2021vit}:
$
\{z^{(t)}\}_{t=1}^{L}, z^{(t)} \in \mathbb{R}^{d}.
$
Our loss can be applied to the tracklet embedding, the per-frame embeddings, or both, by optimizing the sum of the corresponding objectives.
We later present an ablation study to compare these variants.

\section{Experiments}

\noindent\textbf{Datasets and Downstream Setup}\quad
We train and evaluate on publicly available datasets, following the setup proposed in~\cite{parolari2025towards}.
Specifically, we train on 27 videos (85 polyps) from \rc{}~\cite{biffi2024real}. 
It contains 60 full-procedure recordings with frame-level bounding-box annotations and lesion-level metadata, \textit{e.g.}, histology and size. 
We evaluate the generalization of the learned polyp tracklet representations on four downstream tasks.

\textit{Retrieval.}
This task measures the ability to retrieve same-polyp tracklets.
Given a query tracklet, we rank all other tracklets by cosine similarity in the embedding space.
We evaluate retrieval on \rc{} test set (19 videos, 47 polyps) and report performance using mean Average Precision (mAP) and Hit Rate@K (HR@K). A hit occurs if at least one tracklet with the same polyp identity is retrieved within the top-$K$ results.

\textit{ReID.} 
Re-identification measures the ability to decide whether two tracklets depict the same polyp entity or not.
Using frozen embeddings, we compute the similarity between each embedding pair and threshold the scores to classify same-polyp vs different-polyp pairs.
We employ the same evaluation set as in the retrieval task and measure performance with AUROC and AUPR.

\textit{Size estimation.}
It is formulated as binary classification (diminutive, \textit{i.e.} $\leq 5$mm, vs non-diminutive) using non-linear probing.
We freeze the encoder and train a classifier.
The evaluation set combines \rc{}~\cite{biffi2024real} test set with two out-of-distribution datasets, namely \sun{}~\cite{misawa2021development} and \psize{}~\cite{song2025polypsize}, totaling 161 videos (189 polyps). 
We split it by polyp entities, using 70\% for training and 30\% for evaluation.
We report identity-weighted macro F1-Score.

\textit{Histology classification.}
It is treated as binary classification (adenoma vs non-adenoma) with the same non-linear probing setup.
The evaluation set extends that of the size estimation task by additionally including \pset{}~\cite{li2021colonoscopy}, for a total of 195 videos (210 polyps). 
Performance is measured with accuracy. 
\\

\begin{table}[t]
\resizebox{\textwidth}{!}{%
\centering
\setlength{\tabcolsep}{4pt} 
\begin{tabular}{p{2.8cm}cc|ccc|cc|c|c}
\toprule

\multirow{3}{*}{Method}
& & 
& \multicolumn{3}{c|}{\textit{Retrieval}} 
& \multicolumn{2}{c|}{\textit{ReID}} 
& \textit{Size}
& \textit{Histology} \\

& Params & FLOPs 
& mAP & HR@1 & HR@5
& AUROC & AUPR 
& F1-Score
& Acc \\

& (M) & (G) & $(\uparrow)$ & $(\uparrow)$ & $(\uparrow)$ & $(\uparrow)$ & $(\uparrow)$ & $(\uparrow)$ & $(\uparrow)$ \\
\midrule

\multicolumn{3}{l}{\textit{General Purpose FM}} & & & & & & & \\

DinoV2-S/14~\cite{oquab2024dinov2}
& 22.1 & 44.2
& 47.31 & 89.81 & 95.52
& 80.07 & 29.77
& 62.81
& 61.68 \\

DinoV2-B/14~\cite{oquab2024dinov2}
& 86.6 & 175.7
& 46.40 & 87.77 & 95.79
& 78.74 & 26.99
& 67.97
& 66.57 \\

DinoV2-L/14~\cite{oquab2024dinov2}
& 304.4 & 622.5
& 43.08 & 87.50 & 94.97
& 75.64 & 23.77
& 68.02
& 70.63 \\

DinoV2-g/14~\cite{oquab2024dinov2}
& 1136 & 2331
& 48.29 & \textbf{90.35} & 95.92
& 78.23 & 27.04
& \textbf{70.22}
& 70.14 \\

DinoV3-S/16~\cite{simeoni2025dinov3}
& 21.6 & 34.7
& 48.95 & 88.99 & \underline{96.33} 
& 80.47 & 31.77 
& 64.57 
& 67.13 \\

DinoV3-B/16~\cite{simeoni2025dinov3}
& 85.7 & 137.7
& 47.17 & 88.45 & 96.06 
& 77.04 & 26.54 
& \underline{69.67}
& 71.54 \\

DinoV3-L/16~\cite{simeoni2025dinov3}
& 303.1 & 487.2
& 43.89 & 87.91 & 94.97
& 68.97 & 20.54 
& 66.58 
& \underline{75.59} \\

V-JEPA2~\cite{assran2025vjepa2}
& 326.0 & 272.1
& 29.01 & 70.79 & 88.32 
& 69.84 & 17.14 
& 68.68
& 61.54 \\

\hline
\textit{Endoscopy FM} & & & & & & & & & \\

EndoFM~\cite{wang2023endofm}
& 121.3 & 190.1
& 45.75 & 87.64 & 94.70 
& 79.71 & 31.91 
& 75.15*
& 68.32 \\

EndoFM-LV~\cite{wang2025endofmlv}
& 121.3 & 190.1
& 28.21 & 63.18 & 81.39 
& 67.56 & 19.28 
& 61.99
& 68.60 \\

EndoViT~\cite{batic2024endovit}
& 86.6 & 134.9
& 31.12 & 68.34 & 85.33 
& 68.26 & 17.51 
& 56.52
& 61.54 \\

SurgeNet~\cite{jaspers2026surgenet}
& 24.3 & 31.0
& 44.75 & 85.87 & 94.70 
& 75.86 & 25.23 
& 68.83
& 64.62 \\

\hline
\textit{Fully supervised} & & & & & & & & & \\

Sup~\cite{parolari2025temporally}
& 25.6 & 33.1
& 54.72 & 81.93 & 94.97 
& 93.85 & 41.58 
& 62.35
& 72.66 \\

TA~\cite{parolari2025temporally} 
& 25.6 & 33.1
& \underline{57.52} & 84.24 & 95.24 
& \textbf{94.33} & 43.81 
& 63.87
& 73.29 \\

\hline
\hline
\textit{Self-supervised} & & & & & & & & & \\

TPC~\cite{parolari2025towards}
& 25.6 & 33.1
& 43.25 & 75.82 & 91.85
& 78.40 & 27.58
& 64.13
& 64.06 \\

MVE~\cite{intrator2023self}
& 25.6 & 33.1
& 42.05 & 81.25 & 94.16
& 79.73 & 27.47
& 58.82
& 66.22 \\

SFE~\cite{intrator2023self}
& 28.0 & 33.1
& 51.04 & 86.96 & 95.52
& 89.36 & \underline{46.13}
& 65.35 
& 63.22 \\

\rowcolor{orange!20}
\textbf{Ours}
& 25.6 & 33.1
& \textbf{63.13} & \textbf{90.22} & \textbf{96.47}
& \underline{94.17} & \textbf{51.94} 
& \textbf{70.24}
& \textbf{82.38} \\

\bottomrule
\end{tabular}
}
\caption{Comparison with State-of-the-Art. All metrics are in percentage. \textbf{Best}, \underline{second best}; shared bolding denotes no statistical difference ($p>0.05$). *We note that EndoFM~\cite{wang2023endofm} is trained on SUN-SEG~\cite{ji2022sunseg}, a superset of SUN~\cite{misawa2021development} annotated with segmentation masks, leading to an overlap with our test set.}
\label{tab:sota}
\end{table}

\noindent\textbf{Implementation Details}\quad
The encoder is implemented following~\cite{parolari2025temporally}. 
Frame embeddings are extracted with a ResNet-50~\cite{he2016deep} and aggregated by a stack of three transformer encoder layers~\cite{vaswani2017attention} with $8$ heads, $d_{\text{model}} = 256$, $d_{\text{feedforward}} = 1024$ and dropout~$= 0.1$.
Following literature~\cite{chen2020simple}, we project the representations for contrastive learning with an MLP ($d_\text{hidden} = 256$, ReLU, $d_\text{out} = 128$).
The bag size is set to $K = 4$, temperatures to $\tau_{\min} = 0.3$, $\tau_{\max} = 12.0$, batch size to $60$.  
Following~\cite{parolari2025towards,parolari2025temporally}, a tracklet is formed from ground-truth polyp detections whose Intersection Over Union is at least $0.1$ between consecutive frames, subsampled to 1 frame every 4, cropped to $5\times$ the diagonal of the bounding box, and resized to $232\times232$. Tracklet length is set to $L = 8$.
For the non-linear probing setup we use an MLP ($d_\text{hidden} = 256$, GELU~\cite{hendrycks2016gaussian}, dropout = 0.1), trained for 20 epochs using AdamW~\cite{loshchilov2019adamw} with learning rate of $10^{-4}$ for size, $10^{-5}$ for histology. 
Batch size is $64$.
Hardware: 8 CPU cores, 64 GB RAM, NVIDIA A100 (64 GB).
\\

\noindent\textbf{Comparison with State-of-the-Art}\quad
We compare our approach to state-of-the-art methods in Tab.~\ref{tab:sota}.
For fair comparison, fully supervised and self-supervised baselines are re-trained on the same data used in our experiments.
Our method outperforms all prior self-supervised approaches, with 23.69\% relative improvement on retrieval mAP, 5.38\% and 12.60\% on ReID AUROC and AUPR, 7.48\% on size F1-Score and 24.4\% on histology accuracy.
This large margin highlights the effectiveness of the proposed temporal self-supervision and noise-aware loss.
Compared to fully supervised baselines, our method consistently achieves superior performance across tasks.
On ReID, AUROC is comparable to TA~\cite{parolari2025temporally}, while AUPR improves by 18.56\%, indicating higher precision (fewer false positive matches) at comparable recall.
For completeness, we also report results for both general-purpose and endoscopy-specific foundation models. 
The Dino family provides a strong baseline, yet our method performs better overall; we match or surpass DinoV2-giant on all tasks while using less than 3\% of its parameters and being trained on only 27 videos, whereas DinoV2 models are pretrained on hundreds of millions of images. 
This improved efficiency (fewer parameters and FLOPs) also facilitates deployment, making in-procedure support more practical. 
Endoscopy-specific foundation models generally underperform general-purpose ones, consistent with their smaller scale and training data biased toward surgical videos rather than colonoscopy. 
\\

\noindent\textbf{Ablation Study}\quad 
In Tab.~\ref{tab:ablation} (Top) we analyze the impact of the exponential sampling over ranks and curriculum. 
We observe that even without exponential sampling or curriculum, our approach already surpasses the prior self-supervised baselines~\cite{intrator2023self,parolari2025towards}, underscoring the effectiveness of the noise-aware loss at leveraging temporal supervision.
Adding these components yields consistent improvements, with the largest gains on retrieval and re-identification. 
On retrieval, despite comparable HR@1 and HR@5, the 14.95\% mAP increase indicates improved global ranking and a more coherent embedding space.
In Tab.~\ref{tab:ablation} (Bottom) we study the effect of applying the loss at the tracklet level, frame level, or both.
The latter yields the best overall performance. On retrieval, hit rate is slightly higher with tracklet-only embeddings, likely because they more directly emphasizes top-ranked matches rather than overall ranking consistency.
Frame embeddings alone consistently underperform, suggesting noisier supervision.
Nevertheless, all versions outperform self-supervised and supervised baselines.

\begin{table}[t]
\resizebox{\textwidth}{!}{%
\centering
\setlength{\tabcolsep}{4pt} 
\begin{tabular}{cc|ccc|cc|c|c}
\toprule
&
& \multicolumn{3}{c|}{\textit{Retrieval}} 
& \multicolumn{2}{c|}{\textit{ReID}} 
& \textit{Size}
& \textit{Histology} \\

&
& mAP & HR@1 & HR@5 
& AUROC & AUPR 
& F1-Score
& Acc \\
\midrule

Exp. Sampling & Curriculum & & & & & & & \\

\xmark & \xmark 
& 54.92 & 89.40 & \textbf{97.28} 
& 91.92 & 50.53 
& 69.10 
& 80.07 \\

\cmark & \xmark 
& 58.83 & 89.67 & 96.60 
& 92.77 & 48.23 
& 66.43 
& 80.63 \\

\cmark & \cmark
& \textbf{63.13} & \textbf{90.22} & 96.47 
& \textbf{94.17} & \textbf{51.94}
& \textbf{70.24}
& \textbf{82.38} \\

\midrule

Frame & Tracklet & & & & & & & \\

\cmark & \xmark
& 62.15 & 90.90 & 97.42 
& 93.85 & 50.84
& 66.83 
& 80.42 \\

\xmark & \cmark
& 61.49 & \textbf{91.17} & \textbf{97.55} 
& 93.89 & 51.78 
& \textbf{70.20}
& 81.82 \\

\cmark & \cmark
& \textbf{63.13} & 90.22 & 96.47 
& \textbf{94.17} & \textbf{51.94}
& \textbf{70.24}
& \textbf{82.38} \\

\bottomrule
\end{tabular}
}
\caption{Ablation study. (Top) Impact of the exponential rank-based sampling used to construct bags, and curriculum strategy that controls the trade-off between safer and more diverse associations during training. (Bottom) Effect of applying the proposed loss on frame or tracklet embeddings, or both.}
\label{tab:ablation}
\end{table}

\section{Conclusion}

We present a self-supervised framework for tracklet representation learning in colonoscopy.
Self-supervision is derived from temporal structure and optimized with a contrastive loss that mitigates noisy associations.
Extensive experiments demonstrate strong transfer across downstream tasks, outperforming prior self-supervised and supervised baselines and matching or exceeding foundation models with a lightweight encoder while being trained on only 27 videos.
These findings enable scalable representation learning in colonoscopy, reducing reliance on annotations while maintaining performance on clinical applications.

\begin{credits}
\subsubsection{\ackname}
We acknowledge ISCRA for awarding this project access to the LEONARDO supercomputer, owned by the EuroHPC Joint Undertaking, hosted by CINECA (Italy).

\subsubsection{\discintname}
C.B. is affiliated with Cosmo Intelligent Medical Devices, the developer of the GI Genius medical device.
\end{credits}

%
%
%
\bibliographystyle{splncs04}
\bibliography{main}
\end{document}